\documentclass[11pt]{article}

\usepackage[letterpaper,margin=1in]{geometry}
\usepackage{graphicx}
\usepackage{booktabs}
\usepackage{multirow}
\usepackage{amsmath,amssymb}
\usepackage{algorithm}
\usepackage{algorithmic}
\usepackage[font=small,labelfont=bf]{caption}
\usepackage{url}
\usepackage[hidelinks]{hyperref}
\usepackage{microtype}

\title{IDM-Net: A Lightweight Illumination-Decoupled Modulation Network for Low-Light Image Enhancement}

\author{
Cheng-Yen Hsiao \qquad Jing-Ming Guo$^{*}$\\[3pt]
Department of Electrical Engineering\\
National Taiwan University of Science and Technology\\
Taipei, Taiwan\\[3pt]
}
\date{}

\begin{document}

\maketitle

\begin{abstract}
Low-light image enhancement (LLIE) remains challenging for lightweight models because illumination restoration and color fidelity are difficult to optimize simultaneously in the RGB color space. Although recent color-decoupled methods separate luminance and chrominance representations, they primarily optimize luminance as an enhancement target, leaving its potential as an explicit guidance prior largely unexplored during feature reconstruction. To address this limitation, we propose IDM-Net, a lightweight Illumination-Decoupled Modulation Network for low-light image enhancement. IDM-Net adopts a dual-encoder architecture consisting of a structure encoder that extracts multi-scale appearance features from the RGB image and a lightweight illumination encoder that learns illumination priors from the decoupled luminance (Y) channel. To effectively exploit these priors, we introduce an Illumination-Guided Modulation (IGM) module that injects multi-scale illumination cues into the decoder through spatially adaptive affine modulation, enabling accurate brightness restoration while preserving natural color consistency. Furthermore, we design a lightweight Feature Refinement Block (FRB) to progressively suppress degradation artifacts and recover fine-grained image details during reconstruction. Extensive experiments on multiple standard low-light image enhancement benchmarks demonstrate that IDM-Net achieves competitive performance among lightweight LLIE methods while maintaining an excellent balance between restoration quality and computational efficiency.

\end{abstract}


\section{Introduction }

Low-light image enhancement (LLIE) plays a critical role in numerous real-world applications, including mobile photography, intelligent surveillance, autonomous driving, and medical imaging. Images captured under insufficient illumination often suffer from low brightness, severe noise contamination, and color distortion, leading not only to degraded visual quality but also to performance deterioration in downstream vision tasks such as object detection and semantic segmentation. With the increasing demand for deployment on resource-constrained devices, developing LLIE models that simultaneously achieve high restoration quality and computational efficiency has become an important research challenge.

Existing LLIE methods can be broadly categorized into three paradigms. Retinex-based methods \cite{Wei2018RetinexNet,Zhang2019KinD,Cai2023Retinexformer} explicitly model image formation by decomposing an image into illumination and reflectance components for separate enhancement. Although physically interpretable, the decomposition process is inherently illposed and often requires complex multi-branch architectures to ensure stable optimization, limiting their applicability in lightweight scenarios. End-to-end RGB-based methods \cite{Ren2019DeepHybrid,Guo2020ZeroDCE,Liu2021RUAS,Li2022ZeroDCEPP,Ma2022SCI,Wang2023LLFormer} directly learn the mapping from low-light images to normal light images using deep neural networks. While these methods have demonstrated impressive restoration performance, the entanglement of luminance and chrominance information in the RGB color space makes it challenging to simultaneously optimize brightness restoration and color fidelity.

To address this limitation, recent studies have explored color-space-based enhancement frameworks that explicitly disentangle illumination and color information. For instance, LYT-Net \cite{Brateanu2025LYTNet} transforms images into the YUV color space to separately process luminance and chrominance components, while CIDNet \cite{Yan2025HVI} introduces the HVI color space to decouple illumination intensity from color representation. These methods demonstrate that color-space transformation can simplify optimization and improve restoration quality by reducing the interference between brightness and color restoration.
Nevertheless, existing color-decoupled methods primarily treat luminance as an enhancement target, leaving its role as an illumination prior largely unexplored. As a result, illumination cues cannot effectively guide the restoration process, limiting brightness recovery under complex lighting conditions. Motivated by this observation, we propose IDM-Net, which explicitly exploits luminance priors to guide RGB feature reconstruction through illumination-guided modulation.

The main contributions of this work are summarized as follows:

\begin{itemize}
    \item We revisit the role of luminance in lightweight LLIE by exploiting it as an explicit illumination guidance prior for feature reconstruction.
    \item We propose IDM-Net, a lightweight illumination-decoupled modulation network with a dual-encoder architecture and an Illumination-Guided Modulation (IGM) module for illumination-aware restoration.
    \item We introduce a lightweight Feature Refinement Block (FRB) to enhance detail recovery. Extensive experiments demonstrate that IDM-Net achieves highly competitive performance with an effective quality-efficiency trade-off.
\end{itemize}

\section{Related Works}
Recent advances in deep learning have significantly advanced low-light image enhancement (LLIE), leading to the development of diverse enhancement paradigms and network architectures. This section reviews related work on LLIE, illumination and color decoupling strategies, and lightweight enhancement networks.
\subsubsection{Low-light Image Enhancement}
Low-light image enhancement (LLIE) aims to improve image visibility and perceptual quality under insufficient illumination. With the advancement of deep learning, numerous LLIE methods have been proposed to address low contrast, noise amplification, and detail degradation. Early approaches relied on convolutional neural networks (CNNs) to learn direct low-to-normal-light mappings, significantly outperforming traditional techniques. To handle coupled degradations, Bread~\cite{Hu2023Bread} decouples luminance from chrominance and sequentially performs illumination adjustment, noise suppression, and color restoration to alleviate color distortion from brightening, while DDNet~\cite{10.1109/TITS.2024.3359755} adopts a lightweight encoder-decoder that decomposes enhancement into color and gradient subtasks for real-time efficiency.
To further improve restoration quality, researchers explored paradigms beyond CNNs. 

EnlightenGAN~\cite{Jiang2021EnlightenGAN} introduces an unsupervised adversarial framework removing the need for paired data, and LLFlow~\cite{Wang2022LLFlow} formulates enhancement as a distribution transformation via normalizing flow. GLARE~\cite{han2024glare} further derives a codebook prior from normal light images via vector quantization and uses an invertible latent normalizing flow to align low-light features for accurate code retrieval, demonstrating the effectiveness of generative modeling.

Transformer-based architectures have shown strong capability in capturing long-range dependencies. LLFormer~\cite{Wang2023LLFormer} incorporates illumination-aware attention and adaptive exposure fusion, but the quadratic complexity of self-attention incurs substantial overhead. To improve efficiency, MambaLLIE~\cite{Weng2024MambaLLIE} introduces a selective state-space mechanism modeling long-range dependencies with linear complexity. In parallel, diffusion-based methods such as GSAD~\cite{Hou2023GSAD} progressively restore illumination through iterative denoising, showing strong detail recovery and natural reconstruction.
Despite this progress, accurately restoring illumination while preserving natural color and fine structural details remains challenging, especially under complex low-light conditions.

\subsubsection{Illumination Modeling and Color-Decoupled Enhancement}
Illumination modeling plays a crucial role in low-light image enhancement, as insufficient lighting is one of the primary causes of visibility degradation. Inspired by Retinex theory, many LLIE methods explicitly model illumination information to facilitate image restoration. LIME~\cite{Guo2017LIME} estimates illumination maps for brightness enhancement, while RetinexNet \cite{Wei2018RetinexNet}and KinD~\cite{Zhang2019KinD} incorporate Retinex decomposition into deep networks by separating reflectance and illumination components. More recently, Retinexformer~\cite{Cai2023Retinexformer} integrates illumination priors into a transformer framework to better capture global contextual information for low-light enhancement.

Despite the success of illumination modeling, directly enhancing images in the RGB color space remains challenging due to the strong coupling between luminance and chrominance information, which often leads to color distortion. To alleviate this issue, several studies perform enhancement in alternative color spaces. Bread \cite{Hu2023Bread} exploit YCbCr decomposition for illumination enhancement, LYT-Net \cite{Brateanu2025LYTNet} processes luminance and chrominance separately in the YUV space, and Zhou et al. \cite{Zhou02012025} perform enhancement in the HSV domain to better preserve color appearance. These methods demonstrate that color-space decoupling can improve illumination restoration while maintaining color fidelity.

Recent CIDNet \cite{Yan2025HVI} further introduces a learnable color space and adopts a dual-branch architecture to separately restore illumination and chromatic information, highlighting the benefits of luminance–chrominance disentanglement. Nevertheless, existing illumination-aware and color-decoupled methods mainly focus on independent illumination and color modeling, resulting in limited interaction between illumination cues and feature reconstruction. Consequently, illumination information cannot be fully exploited throughout the restoration process. To address this limitation, we propose IDM-Net, which employs a dedicated illumination encoder and an Illumination-Guided Modulation (IGM) module to inject multi-scale illumination priors into RGB feature decoding. By explicitly guiding feature reconstruction with illumination information, IDM-Net achieves accurate brightness restoration while preserving natural color appearance.

\begin{figure*}[t] 
\centering
\includegraphics[width=\textwidth]{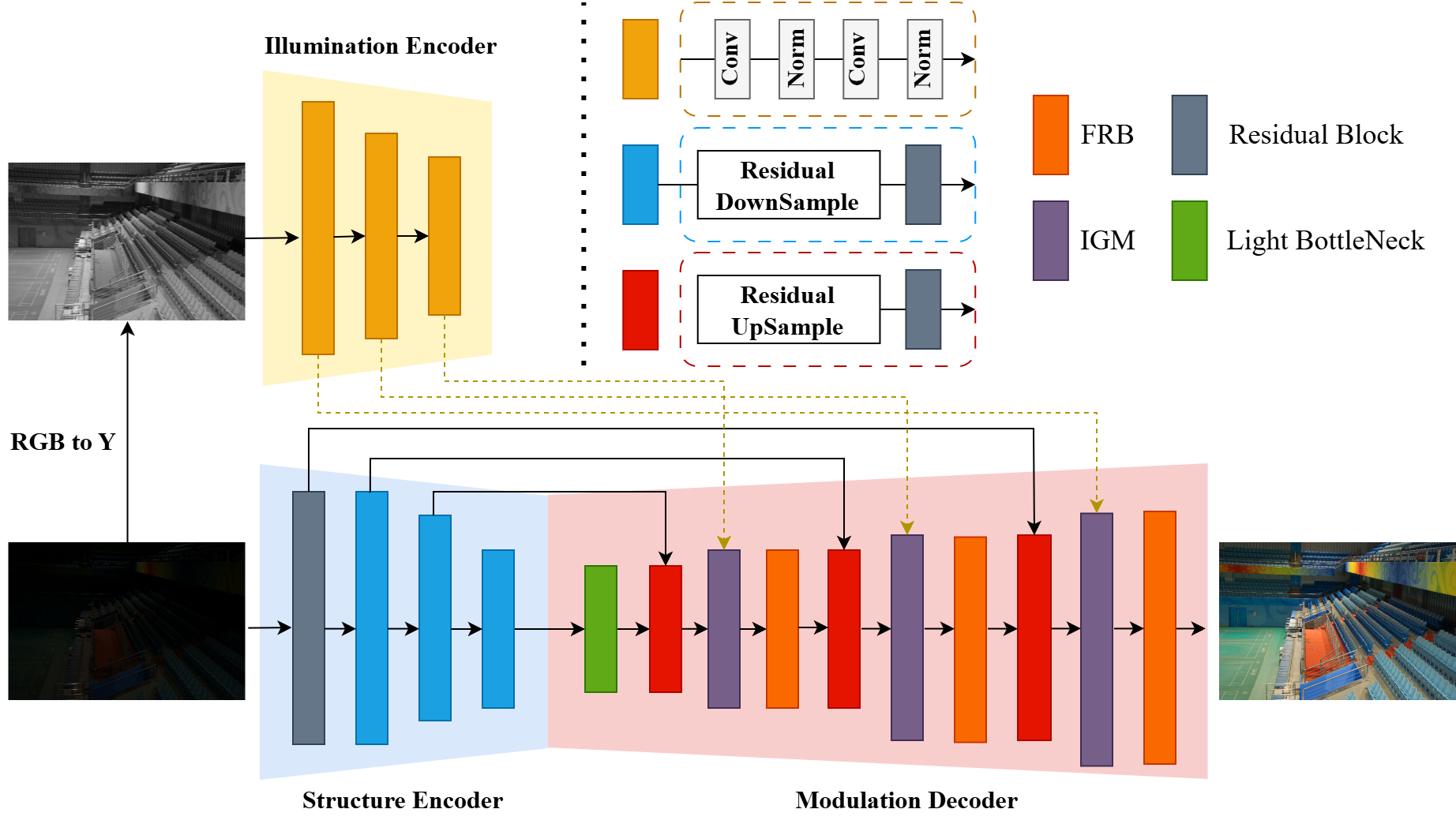} 
\caption{Overview of IDM-Net. Instead of directly restoring luminance, the proposed framework exploits luminance features as illumination priors to guide multi-scale RGB feature reconstruction.}
\label{fig:arc}
\end{figure*}

\subsubsection{Lightweight LLIE Networks}
Although recent LLIE methods have achieved impressive performance through illumination modeling, color-space decomposition, and advanced network architectures, many approaches rely on increasingly complex designs, resulting in high computational and memory costs. This limits their deployment on resource-constrained and real-time applications. Consequently, developing lightweight LLIE networks that maintain enhancement quality while reducing computational complexity has become an important research direction.

Existing lightweight LLIE methods address this challenge from different perspectives. Zero-DCE \cite{Guo2020ZeroDCE} formulates enhancement as a curve estimation problem with non-reference learning, while RUAS \cite{Liu2021RUAS} incorporates illumination priors into a lightweight architecture discovered through neural architecture search. IAT \cite{Cui2022IAT} combines lightweight local modeling and global adjustments for efficient enhancement, whereas PairLIE \cite{Fu2023PairLIE} and Multinex \cite{Brateanu2026Multinex} improve representation capability through self-supervised learning and efficient luminance-reflectance decomposition, respectively. In addition, CPGA-Net \cite{Weng2025CPGANet} and CPGA-Net+ \cite{Weng2025RTI} integrate handcrafted priors to enhance restoration quality within compact network architectures.

Despite these advances, lightweight LLIE methods often face a trade-off between computational efficiency and restoration quality. Accurately restoring illumination while preserving color consistency and fine details remains challenging under limited computational budgets. Moreover, illumination information is often processed independently or insufficiently integrated into feature reconstruction, restricting effective illumination-aware enhancement. To address these limitations, we propose IDM-Net, which introduces illumination-guided feature modulation and lightweight feature refinement for efficient and color-consistent low-light image enhancement.

\section{Methodology}
As shown in Figure \ref{fig:arc}, IDM-Net employs a dual encoder–decoder architecture to separately learn structural and illumination features. Illumination priors guide structural restoration through the Illumination-Guided Modulation (IGM) module, while the Feature Refinement Block (FRB) enhances feature representation and detail recovery. This design enables effective low-light image enhancement with low computational cost.

\subsubsection{The Illumination-Decoupled Dual Encoder}
To explicitly decouple illumination from appearance information, IDM-Net adopts a dual-encoder architecture that processes the input RGB image and its luminance component in parallel. This design is motivated by the observation that enhancing illumination directly in the RGB space often alters the original color relationships, leading to color shifts and visual artifacts. By separating illumination modeling from appearance representation, our network can restore brightness without distorting structural and chromatic information.

Given a low-light RGB image $I_{in} \in \mathbb{R}^{3 \times H \times W}$, we first extract its luminance component $Y \in \mathbb{R}^{1 \times H \times W}$ via the standard BT.601 luma transform:
\begin{equation}
Y = 0.299 \cdot R + 0.587 \cdot G + 0.114 \cdot B
\end{equation}
which provides an explicit, noise-robust representation of scene illumination. The original RGB image and the extracted luminance channel are then fed into two parallel encoders. The Structural Encoder extracts multi-scale appearance features that preserve structural details and chromatic information, while the Illumination Encoder focuses on learning illumination-aware representations from the decoupled luminance component.
Formally, the two encoders progressively extract multi-scale feature representations through sequential downsampling:
\begin{align}
F_S^l &= \varepsilon_S^l (F_S^{l-1}), \quad F_S^0 = I_{in}, \\
F_Y^l &= \varepsilon_Y^l (F_Y^{l-1}), \quad F_Y^0 = Y,
\end{align}
For $l \in \{1, 2, \dots, L\}$, where $\varepsilon_S^l$ and $\varepsilon_Y^l$ denote the $l$-th encoder block of the Structural Encoder and Illumination Encoder, respectively. After $L$ stages, the dual encoder produces two complementary sets of multi-scale features: $\{F_S^l\}_{l=1}^L$ encoding structural and chromatic information, and $\{F_Y^l\}_{l=1}^L$ encoding illumination priors at corresponding scales. These features are subsequently utilized by the decoder for illumination-guided reconstruction.

Both encoders are constructed using lightweight convolutional Residual Blocks followed by strided downsampling operations. The Residual Blocks enhance feature extraction capability while maintaining efficient information propagation, whereas the downsampling layers progressively reduce spatial resolution to capture higher-level contextual information. Through multi-scale feature extraction, the Structural Encoder focuses on preserving scene structures and edge details, while the Illumination Encoder learns illumination-aware representations for subsequent brightness restoration.

Unlike conventional RGB-only feature extraction, the proposed dual-encoder architecture generates two complementary representations. The structural features preserve scene content and texture details, while the illumination features provide explicit illumination priors that are less affected by chromatic variations. Such decoupling facilitates more effective illumination-aware reconstruction in subsequent decoding stages and helps alleviate color distortions commonly observed in RGB-based enhancement methods.

\subsubsection{Illumination-Guided Modulation (IGM)}
While simple fusion strategies such as concatenation or element-wise addition can combine structural and illumination features, they treat illumination priors as ordinary feature responses and provide only passive guidance. Since illumination priors may gradually dilute after successive convolution operations, such implicit fusion is insufficient for spatially non-uniform brightness restoration. To address this, we propose the Illumination-Guided Modulation (IGM) module, which explicitly injects illumination priors into the reconstruction process through spatially adaptive affine transformation.

Given the structural feature $F_S^l$ and the corresponding illumination feature $F_Y^l$ at the $l$-th decoder scale, the illumination feature is first resized to match the spatial resolution of $F_S^l$ when necessary. Since illumination degradation is typically spatially non-uniform, directly estimating modulation parameters from the entire illumination feature map may introduce irrelevant responses from well-exposed regions. A spatial attention mechanism is therefore applied to emphasize illumination-sensitive regions:
\begin{equation}
A^l = \sigma \left( \text{Conv}_{7 \times 7} \left( F_Y^l \right) \right)
\end{equation}
where $\sigma(\cdot)$ denotes the sigmoid activation. A large-kernel convolution ($7 \times 7$) is adopted as illumination transitions tend to be spatially smooth and span larger spatial regions. The attention map recalibrates the illumination feature as:
\begin{equation}
\tilde{F}_Y^l = A^l \odot F_Y^l
\end{equation}
where $\odot$ denotes element-wise multiplication. Through this operation, the network focuses on regions requiring illumination correction while suppressing less informative responses.

To capture broader illumination context, $\tilde{F}_Y^l$ is subsequently processed by an illumination projection network comprising a standard $3 \times 3$ convolution followed by a dilated convolution (dilation rate $d$), which expands the effective receptive field to model large-scale illumination variations such as global exposure differences and spatial lighting gradients. The projection network generates two spatially adaptive affine parameters:
\begin{equation}
[\gamma^l, \beta^l] = \phi \left( \tilde{F}_Y^l \right)
\end{equation}
where $\phi(\cdot)$ denotes the learnable projection function, and $\gamma^l, \beta^l$ represent the scale and shift parameters, respectively.

Finally, the affine parameters modulate the structural feature in a SPADE-like spatially adaptive manner~\cite{Park2019SPADE}:
\begin{equation}
\tilde{F}_S^l = \text{GN}(F_S^l) \odot (1 + \gamma^l) + \beta^l
\end{equation}
where $\text{GN}(\cdot)$ denotes Group Normalization. Here, $\gamma^l$ adaptively controls the response intensity of structural features according to local illumination conditions, while $\beta^l$ provides illumination-dependent bias compensation.

By replacing passive feature fusion with explicit adaptive modulation, IGM ensures that illumination priors persistently guide structural feature reconstruction at each decoder scale. The combination of spatial attention, which highlights illumination-sensitive regions, and dilated context modeling, which captures large-scale lighting distributions, enables the network to perform spatially adaptive brightness restoration while preserving chromatic consistency.

\begin{figure*}[t] 
\centering
\includegraphics[width=0.9\textwidth]{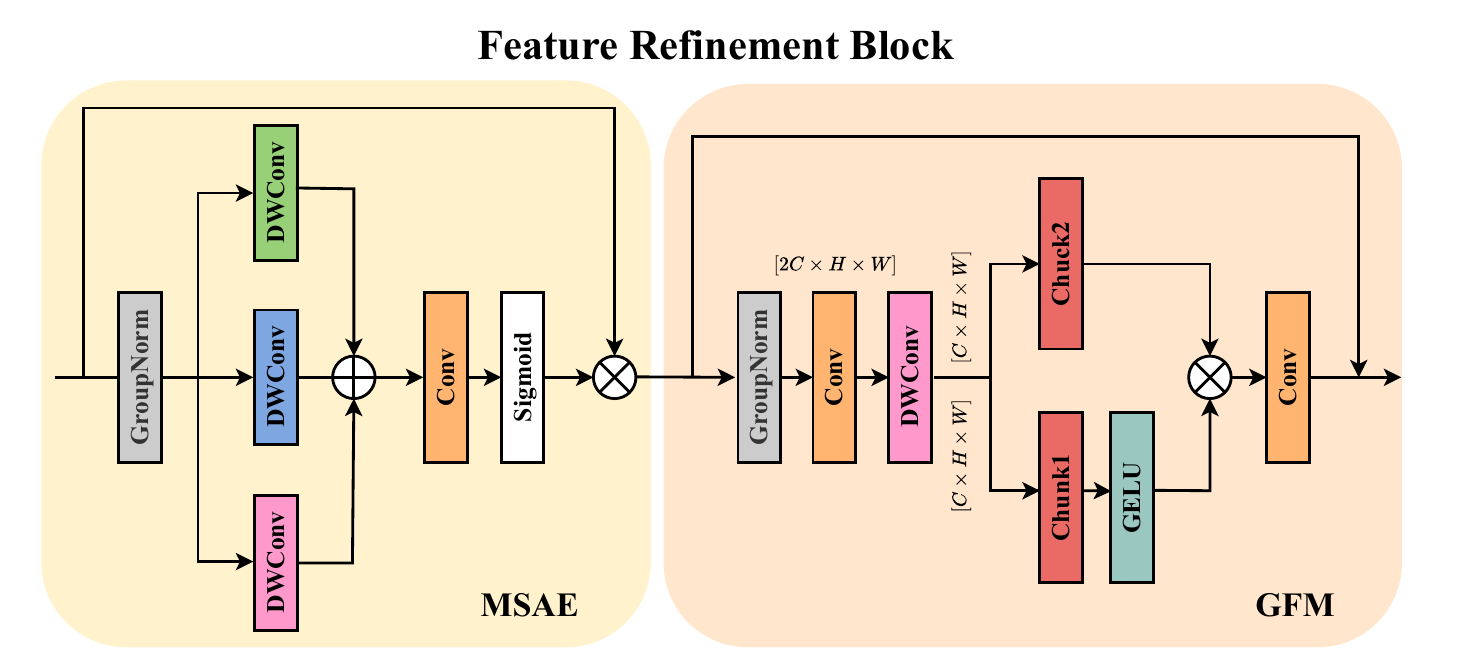} 
\caption{Architecture of the proposed Feature Refinement Block (FRB), consisting of a Multi-Scale Adaptive Excitation (MSAE) module and a Gated Feature Modulation (GFM) module.}
\label{fig:FRB}
\end{figure*}

\subsubsection{Feature Refinement Block (FRB)}
Although the IGM module effectively injects illumination priors into the reconstruction process, the decoder features may still contain residual noise and structural artifacts introduced during the low-light capture process. To address this, we propose the Feature Refinement Block (FRB), which progressively suppresses noise while preserving fine-grained structural details throughout the decoding stages. As show in Figure~\ref{fig:FRB}, FRB consists of two complementary components: a Multi-Scale Adaptive Excitation (MSAE) module for multi-scale noise localization, and a Gated Feature Modulation (GFM) for selective feature refinement.

Multi-Scale Adaptive Excitation (MSAE): Noise in low-light images exhibits spatially irregular distributions that span multiple scales, making it difficult to suppress with a fixed receptive field. To capture noise patterns at different scales simultaneously, MSAE employs three parallel depthwise dilated convolutions with different dilation rates:
\begin{equation}
x_i = \text{DWConv}_{d_i} (x), \quad i \in \{1, 2, 3\},
\end{equation}
where $\text{DWConv}_{d_i}(\cdot)$ denotes a depthwise convolution with dilation rate $d_i$. The multi-scale responses are concatenated and projected through a $1 \times 1$ convolution to generate a spatial attention map:
\begin{equation}
F = \sigma \left( \text{Conv}_{1 \times 1} \left( [x_1; x_2; x_3] \right) \right)
\end{equation}
where $\sigma(\cdot)$ denotes the sigmoid activation and $[\cdot; \cdot]$ denotes channel-wise concatenation. The attention map $F$ recalibrates the input feature:
\begin{equation}
x = x \odot F
\end{equation}
where $\odot$ denotes element-wise multiplication. Through this operation, MSAE suppresses noisy responses while preserving spatially coherent structural features.

The dilation rates are dataset-dependent: for real-world datasets (LOL-v1 and LOL-v2-real), we adopt $\{d_1, d_2, d_3\} = \{1, 2, 4\}$ to capture fine-grained local noise patterns; for the synthetic dataset (LOL-v2-synthetic), we use $\{1, 4, 9\}$ to model larger-scale structured noise patterns, reflecting the different noise characteristics between real and synthetic data.

Gated Feature Modulation (GFM): After spatial noise suppression by MSAE, GFM further refines the features through a gating mechanism that selectively amplifies informative responses while suppressing redundant ones. Given the input feature $x$, GFM first expands the channel dimension via a $1 \times 1$ convolution followed by a depthwise convolution, then splits the output into two equal halves along the channel dimension. A gating operation is applied by element-wise multiplying the GELU-activated first half with the second half, allowing the network to selectively pass informative activations while suppressing noise-corrupted ones. The gated output is finally projected back to the original channel dimension through a $1 \times 1$ convolution. This design enables expressive feature selection without introducing additional parameters beyond the projection layers.

\begin{figure*}[t]
\centering
\includegraphics[width=\textwidth]{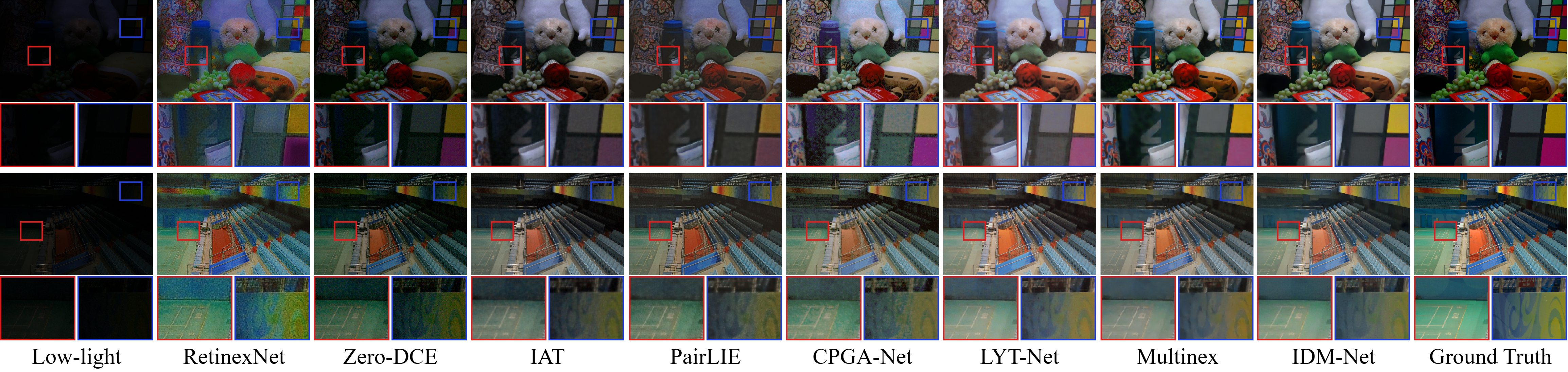} 
\caption{Visual comparison on the LOL datasets. Zoomed-in regions highlight the superior detail restoration and texture preservation achieved by the proposed IDM-Net.}
\label{fig:lol}
\end{figure*}

\section{Experimental Results}
This section evaluates the proposed method on paired and unpaired low-light image enhancement benchmarks. We compare it with state-of-the-art methods in terms of quantitative performance, visual quality, and computational efficiency.

\subsubsection{Datasets and Evaluation Metrics}
Experiments are conducted on both paired and unpaired low-light image enhancement benchmarks. For paired evaluation, LOLv1, LOLv2-real, and LOLv2-synthetic are adopted, while LIME~\cite{Guo2017LIME}, MEF~\cite{Ma2015MEFSSIM}, NPE~\cite{Wang2013NPEA}, VV~\cite{Vonikakis2018Evaluation}, and DICM~\cite{Lee2013LDR} are used for unpaired evaluation. As ground-truth references are unavailable for the unpaired datasets, performance is evaluated using the no-reference NIQE metric, which assesses the naturalness and perceptual quality of the enhanced results.

\subsubsection{Implementation Details}
All experiments are conducted on a workstation equipped with a single NVIDIA GeForce RTX 5060 Ti GPU. The proposed IDM-Net employs a base channel width of 16, resulting in only 0.79M parameters and 5.47 GFLOPs. The LightBottleneck module is implemented using depth-wise separable convolutions, which significantly reduce the parameter count and computational cost while preserving feature representation capability. Unless otherwise specified, three Feature Refinement Blocks (FRBs) are stacked at each decoder scale.

During training, images are used at their original resolutions with random flipping and rotation for data augmentation. The network is trained for 800 epochs using AdamW with an initial learning rate of $2 \times 10^{-4}$, followed by cosine annealing and a 10-epoch warm-up. The batch size is set to 4 for LOL-v1 and LOL-v2-Real, and 1 for LOL-v2-Synthetic. Following standard LLIE protocols, PSNR, SSIM, and LPIPS are adopted for evaluation.

\begin{table*}[t]
\caption{Comparison with state-of-the-art methods on LOL benchmarks. The best and second-best results are highlighted in \textbf{bold} and \underline{underlined}, respectively.}
\label{tab:sota_lol_all_two_lines}
\centering
\small
\setlength{\tabcolsep}{2pt}
\begin{tabular*}{\textwidth}{@{\extracolsep{\fill}}llccccccccc}
\toprule
\multirow{2}{*}{Size} & \multirow{2}{*}{Method}
& \multicolumn{3}{c}{LOLv1}
& \multicolumn{2}{c}{LOLv2-real}
& \multicolumn{2}{c}{LOLv2-syn}
& \multicolumn{2}{c}{Efficiency} \\
\cmidrule(lr){3-5}
\cmidrule(lr){6-7}
\cmidrule(lr){8-9}
\cmidrule(lr){10-11}
& & PSNR$\uparrow$ & SSIM$\uparrow$ & LPIPS$\downarrow$
& PSNR$\uparrow$ & SSIM$\uparrow$
& PSNR$\uparrow$ & SSIM$\uparrow$
& \#P(M)$\downarrow$ & GFLOPs$\downarrow$ \\
\midrule

\multirow{5}{*}{\rotatebox{90}{\parbox{1.4cm}{\centering Heavy\\($>10$M)}}}
& EnGAN    & 17.54 & 0.664 & 0.326 & 18.23 & 0.617 & 16.49 & 0.771 & 54.41 & 22.19 \\
& LLFormer & \textbf{23.65} & 0.816 & 0.169 & \textbf{27.75} & \underline{0.860} & 17.16 & 0.784 & 24.55 & 39.05 \\
& GLARE    & \underline{23.55} & \textbf{0.863} & \textbf{0.086} & 22.51 & \textbf{0.871} & 18.21 & 0.842 & 59.48 & 508.42 \\
& LLFlow   & 21.15 & 0.854 & 0.116 & 17.43 & 0.831 & \textbf{24.81} & 0.919 & 17.42 & 358.40 \\
& GSAD     & 22.77 & 0.852 & \underline{0.102} & 20.15 & 0.846 & \underline{24.47} & \textbf{0.929} & 17.36 & 442.02 \\

\midrule
\multirow{5}{*}{\rotatebox{90}{\parbox{1.4cm}{\centering Mid Size\\($1$--$10$M)}}}
& KinD          & 17.65 & 0.771 & 0.175 & 14.74 & 0.641 & 17.28 & 0.758 & 8.02 & 63.68 \\
& DDNet         & 21.82 & 0.798 & 0.186 & 23.02 & 0.834 & 24.63 & 0.917 & 5.39 & 111.47 \\
& Bread         & 22.96 & 0.838 & 0.155 & 20.83 & 0.847 & 17.63 & 0.919 & 2.02 & 19.85 \\
& CIDNet        & 23.81 & \textbf{0.857} & \textbf{0.086} & \textbf{24.11} & \textbf{0.871} & \textbf{25.71} & \textbf{0.942} & 1.88 & 7.57 \\
& Retinexformer & \textbf{25.15} & \underline{0.846} & \underline{0.131} & \underline{22.79} & \underline{0.840} & \underline{25.67} & \underline{0.930} & 1.53 & 15.85 \\

\midrule
\multirow{9}{*}{\rotatebox{90}{\parbox{1.8cm}{\centering Lightweight\\($<1$M)}}}
& Retinex-Net & 16.77 & 0.425 & 0.474 & 18.37 & 0.723 & 17.14 & 0.756 & 0.56 & 79.61 \\
& PairLIE     & 19.56 & 0.730 & 0.248 & 19.89 & 0.778 & 19.07 & 0.794 & 0.34 & 81.84 \\
& IAT         & 23.38 & 0.809 & 0.210 & \textbf{23.50} & 0.824 & 15.37 & 0.710 & 0.09 & 5.27 \\
& Zero-DCE    & 14.86 & 0.562 & 0.335 & 14.32 & 0.511 & 17.76 & 0.814 & 0.08 & 4.83 \\
& LYT-Net     & 22.38 & 0.826 & \underline{0.134} & 20.97 & 0.840 & 23.50 & 0.914 & 0.05 & 8.04 \\
& CPGA-Net    & 20.94 & 0.748 & 0.260 & 20.79 & 0.759 & 20.68 & 0.833 & 0.03 & 6.03 \\
& CPGA-Net+   & 22.53 & 0.812 & 0.205 & 20.90 & 0.800 & 23.07 & 0.907 & 0.06 & 9.36 \\
& Multinex    & 23.19 & 0.843 & \textbf{0.129} & \underline{23.04} & \textbf{0.860} & \underline{25.04} & \underline{0.930} & 0.05 & 2.50 \\
& \textbf{IDM-Net}     & \textbf{24.15} & \textbf{0.845} & \textbf{0.129} & 23.01 & \underline{0.858} & \textbf{25.11} & \textbf{0.935} & 0.79 & 5.48 \\

\bottomrule
\end{tabular*}
\end{table*}

\begin{table}[t]
\caption{Image quality comparison on unpaired datasets using the NIQE metric, where lower values indicate better performance. The best and second-best results are highlighted in \textbf{bold} and \underline{underlined}, respectively.}
\label{tab:unpaired_niqe}
\centering
\small
\begin{tabular*}{\columnwidth}{@{\extracolsep{\fill}}lcccccc}
\toprule
\multirow{2}{*}{Method} & \multicolumn{6}{c}{Unpaired Datasets (NIQE$\downarrow$)} \\
\cmidrule(lr){2-7}
& MEF & LIME & NPE & VV & DICM & Avg \\
\midrule
Retinex-Net & 4.081 & 4.907 & 4.457 & 2.619 & 4.737 & 4.160 \\
PairLIE     & 4.357 & 4.312 & 4.093 & 3.155 & 3.240 & 3.831 \\
IAT         & 4.169 & 4.174 & \underline{3.290} & 3.174 & 3.032 & 3.568 \\
Zero-DCE    & \textbf{3.507} & \textbf{3.791} & 3.508 & \underline{2.752} & \underline{3.102} & \underline{3.332} \\
\textbf{IDM-Net}     & \underline{3.651} & \underline{3.966} & \textbf{3.152} & \textbf{2.452} & \textbf{2.732} & \textbf{3.191} \\
\bottomrule
\end{tabular*}
\end{table}

\begin{table}[t]
\caption{Ablation study of different components in the proposed network. The best and second-best results are highlighted in \textbf{bold} and \underline{underlined}, respectively.}
\label{tab:ablation_clean}
\centering
\small
\begin{tabular*}{\columnwidth}{@{\extracolsep{\fill}}llccc}
\toprule
& Combination & PSNR$\uparrow$ & SSIM$\uparrow$ & LPIPS$\downarrow$ \\
\midrule
(a) & Base UNet                    & 21.45 & 0.798 & 0.199 \\
(b) & (a) + LightBottleck         & 22.06 & 0.812 & 0.169 \\
(c) & (b) + Y branch + concat     & 23.26 & 0.836 & 0.133 \\
(d) & (b) + Y branch + IGM        & \underline{23.86} & \underline{0.840} & \underline{0.130} \\
(e) & (d) + FRB (Full IDM-Net)    & \textbf{24.15} & \textbf{0.845} & \textbf{0.129} \\
\bottomrule
\end{tabular*}
\end{table}

\subsubsection{Quantitative Results}
Table~\ref{tab:sota_lol_all_two_lines} reports the quantitative results on LOL-v1, LOL-v2-Real, and LOL-v2-Synthetic, while Figure 3 provides visual comparisons with representative LLIE methods. For visual comparison, we select several state-of-the-art lightweight approaches as baselines. As highlighted in the zoomed-in regions of Figure~\ref{fig:lol}, IDM-Net recovers more faithful colors and finer structural details than competing methods. In particular, our method better preserves texture information while avoiding color distortion and over-enhancement artifacts, resulting in more natural and visually pleasing restoration results.

Among lightweight methods, IDM-Net achieves the best overall performance. On LOL-v1, IDM-Net attains 24.15 dB PSNR and 0.845 SSIM, outperforming all other lightweight approaches and approaching the performance of substantially larger models. On LOL-v2-Real, IDM-Net achieves competitive results with 23.01 dB PSNR and 0.858 SSIM, ranking among the top lightweight methods. The relatively smaller improvement on this dataset may be attributed to real sensor noise and photometric inconsistencies. Since IDM-Net utilizes the luminance map derived from the Y channel of the input image as an illumination prior, noise-corrupted luminance information may reduce the reliability of illumination-guided modulation and consequently limit enhancement performance. Nevertheless, IDM-Net maintains favorable reconstruction quality while preserving its lightweight design. On LOL-v2-Synthetic, IDM-Net achieves the best performance among lightweight methods, reaching 25.11 dB PSNR and 0.935 SSIM. Furthermore, IDM-Net consistently delivers strong perceptual quality, achieving one of the lowest LPIPS scores on LOL-v1, demonstrating its ability to generate visually pleasing and perceptually faithful enhancement results.

In terms of efficiency, IDM-Net requires only 0.79M parameters and 5.48 GFLOPs, which is substantially lower than most heavy and mid-size models while maintaining competitive enhancement performance. Compared with representative lightweight approaches such as LYT-Net, CPGA-Net+, and Multinex, IDM-Net provides a better balance between restoration quality and computational complexity, demonstrating the effectiveness of the proposed illumination-decoupled modulation framework.
 
Table~\ref{tab:unpaired_niqe} reports the NIQE results on five unpaired low-light datasets, including MEF, LIME, NPE, VV, and DICM. Lower NIQE values indicate better perceptual image quality.
\begin{figure}[t]
    \centering
    \includegraphics[width=\columnwidth]{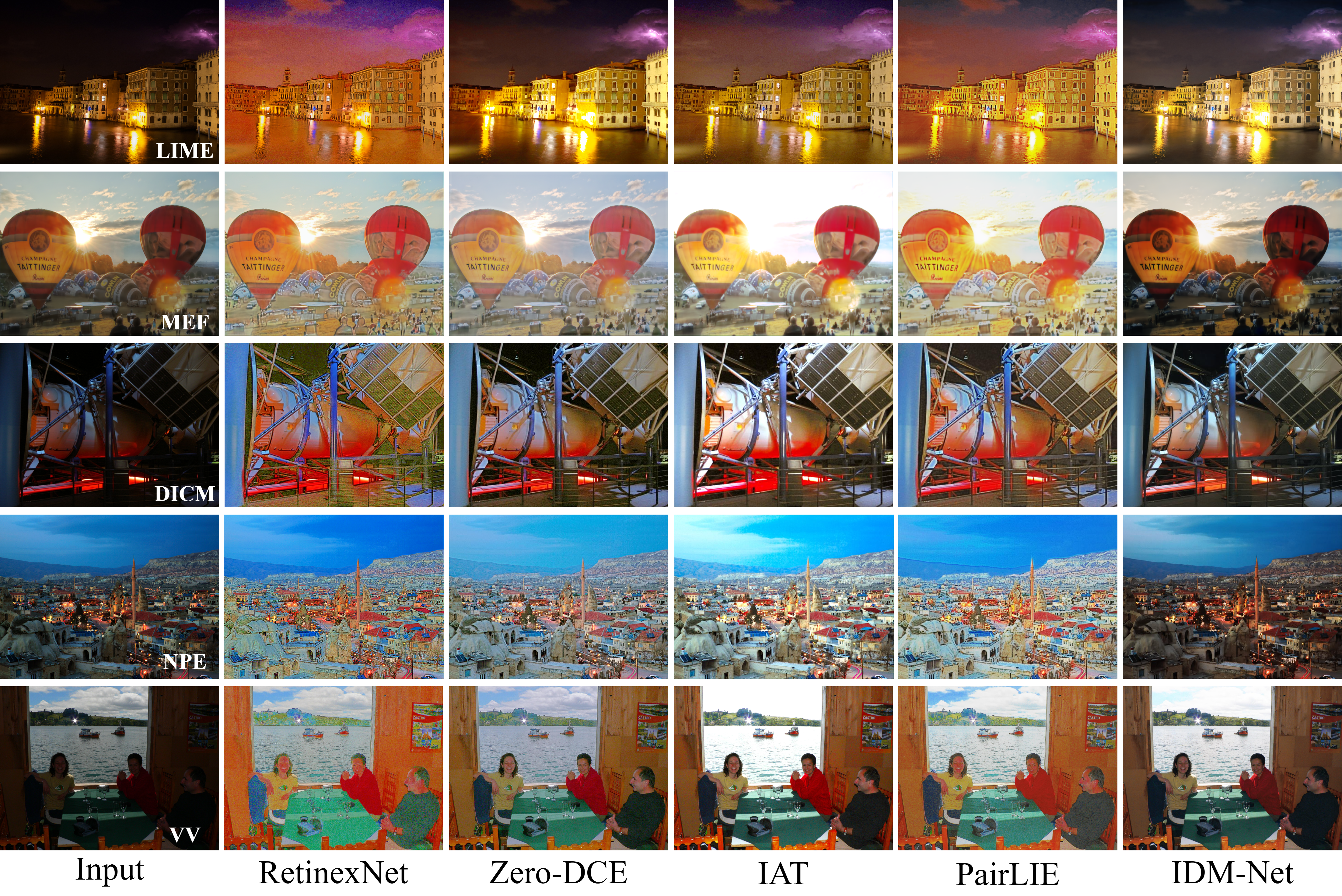}
    \caption{Qualitative comparison of different low-light image
    enhancement methods on representative images from the unpaired
    low-light datasets.}
    \label{fig:unpaired_visual}
\end{figure}

As shown in Table~\ref{tab:unpaired_niqe}, IDM-Net achieves the best average NIQE score of 3.191 among all compared lightweight methods. Specifically, IDM-Net obtains the best results on NPE, VV, and DICM, and achieves the second-best performance on MEF and LIME. These results indicate that the proposed illumination-decoupled design and illumination-guided modulation strategy generalize well beyond paired training datasets and can effectively enhance real-world low-light images. Furthermore, as shown in Fig.~\ref{fig:unpaired_visual}, IDM-Net generally produces more natural restoration results compared with the other methods, with balanced illumination, realistic color reproduction, and well-preserved image details. The superior average NIQE score, together with the qualitative comparison, further demonstrates that IDM-Net is capable of producing visually natural and perceptually pleasing enhancement results under diverse imaging conditions.

\subsubsection{Ablation Study}
To evaluate the contribution of each component in the proposed Network, ablation studies are conducted on the LOL-v1 dataset. Starting from a baseline network, components are progressively introduced to analyze their individual and cumulative effects on restoration performance. The quantitative results are summarized in Table~\ref{tab:ablation_clean}. The baseline model achieves a PSNR of 21.45 dB, an SSIM of 0.798, and an LPIPS score of 0.199. Replacing the standard convolution blocks with the proposed LightBottleneck, which is built upon depthwise convolutions (DWConv) combination, improves the performance to 22.06 dB PSNR and 0.812 SSIM while reducing LPIPS to 0.169. These results demonstrate that the lightweight DWConv-based design effectively enhances feature representation while maintaining low computational complexity.

Building upon the LightBottleneck backbone, introducing the illumination branch and directly fusing illumination features through concatenation further improves the performance to 23.26 dB PSNR and 0.836 SSIM. The significant gain demonstrates the importance of explicitly incorporating illumination information into the enhancement process.

To verify the effectiveness of the proposed Illumination-Guided Modulation (IGM), the concatenation operation is replaced by the proposed adaptive modulation mechanism. As shown in Table~\ref{tab:ablation_clean}, the PSNR increases from 23.26 dB to 23.86 dB, while SSIM improves from 0.836 to 0.840 and LPIPS decreases from 0.133 to 0.130. These results suggest that explicitly modulating structural features with illumination priors is more effective than passive feature fusion, enabling the network to better adapt to spatially varying illumination conditions.

Finally, incorporating the Feature Refinement Block (FRB) leads to the full IDM-Net and further boosts performance to 24.15 dB PSNR, 0.845 SSIM, and 0.129 LPIPS. Compared with the baseline, the complete model achieves gains of 2.70 dB in PSNR and 0.047 in SSIM while reducing LPIPS by 0.07. These results demonstrate that each proposed component contributes positively to the overall performance, and their combination yields the best restoration quality.

\section{Conclusion, Limitation and Future Works}
In this paper, we present IDM-Net, a lightweight Illumination-Decoupled Modulation Network for low-light image enhancement. Rather than relying on explicit illumination-reflectance decomposition as in conventional Retinex-based approaches, IDM-Net reformulates illumination guidance as a feature modulation process: a dedicated Illumination Encoder extracts multi-scale luminance priors, which guide the Structure Encoder's content features through the proposed Illumination-Guided Modulation (IGM) module. Combined with the Feature Refinement Block (FRB), this architectural decoupling enables effective brightness restoration while preserving structural and chromatic fidelity.

Experimental results on LLIE benchmark datasets demonstrate that IDM-Net achieves highly competitive performance among lightweight LLIE methods. Nevertheless, the model may still be affected by photometric inconsistencies in real-world datasets, and the fixed illumination prior may not fully capture the complex characteristics of diverse camera imaging pipelines, which could limit its generalization capability in certain scenarios.

In future work, we plan to investigate learnable illumination priors to further improve enhancement quality and robustness, and explore an ultra-lightweight version of IDM-Net for real-time deployment on resource-constrained devices.

\bibliographystyle{IEEEtran}
\bibliography{cite}

@inproceedings{Wei2018RetinexNet,
  title={Deep Retinex Decomposition for Low-Light Enhancement},
  author={Wei, Chen and Wang, Wenjing and Yang, Wenhan and Liu, Jiaying},
  booktitle={British Machine Vision Conference (BMVC)},
  year={2018}
}

@inproceedings{Zhang2019KinD,
  title={Kindling the Darkness: A Practical Low-Light Image Enhancer},
  author={Zhang, Yonghua and Zhang, Jiawan and Guo, Xiaojie},
  booktitle={Proceedings of the 27th ACM International Conference on Multimedia},
  pages={1632--1640},
  year={2019}
}

@inproceedings{Cai2023Retinexformer,
  title={Retinexformer: One-stage Retinex-based Transformer for Low-Light Image Enhancement},
  author={Cai, Yuanhao and Bian, Hao and Lin, Jing and Wang, Haoqian and Timofte, Radu and Van Gool, Luc},
  booktitle={Proceedings of the IEEE/CVF International Conference on Computer Vision (ICCV)},
  pages={12504--12513},
  year={2023}
}

@article{Ren2019DeepHybrid,
  title={Low-Light Image Enhancement via a Deep Hybrid Network},
  author={Ren, Wenqi and Liu, Si and Ma, Lin and Xu, Qian and Xu, Xiaochun and Cao, Xiaochun and Du, Jun and Yang, Ming-Hsuan},
  journal={IEEE Transactions on Image Processing},
  volume={28},
  number={9},
  pages={4364--4375},
  year={2019}
}

@inproceedings{Guo2020ZeroDCE,
  title={Zero-Reference Deep Curve Estimation for Low-Light Image Enhancement},
  author={Guo, Chunle and Li, Chongyi and Guo, Jichang and Loy, Chen Change and Hou, Junhui and Kwong, Sam and Cong, Runmin},
  booktitle={Proceedings of the IEEE/CVF Conference on Computer Vision and Pattern Recognition (CVPR)},
  pages={1780--1789},
  year={2020}
}

@article{Li2022ZeroDCEPP,
  title={Learning to Enhance Low-Light Image via Zero-Reference Deep Curve Estimation},
  author={Li, Chongyi and Guo, Chunle and Loy, Chen Change},
  journal={IEEE Transactions on Pattern Analysis and Machine Intelligence},
  volume={44},
  number={8},
  pages={4225--4238},
  year={2022}
}

@inproceedings{Liu2021RUAS,
  title={Retinex-inspired Unrolling with Cooperative Prior Architecture Search for Low-Light Image Enhancement},
  author={Liu, Risheng and Ma, Long and Zhang, Jiaao and Fan, Xin and Luo, Zhongxuan},
  booktitle={Proceedings of the IEEE/CVF Conference on Computer Vision and Pattern Recognition (CVPR)},
  pages={10561--10570},
  year={2021}
}

@inproceedings{Ma2022SCI,
  title={Toward Fast, Flexible, and Robust Low-Light Image Enhancement},
  author={Ma, Long and Ma, Tengyu and Liu, Risheng and Fan, Xin and Luo, Zhongxuan},
  booktitle={Proceedings of the IEEE/CVF Conference on Computer Vision and Pattern Recognition (CVPR)},
  pages={5637--5646},
  year={2022}
}

@inproceedings{Wang2023LLFormer,
  title={LLFormer: A Transformer Architecture for Low-Light Image Enhancement},
  author={Wang, Tao and Zhang, Kai and Shen, Wenhan and Luo, Wenhan and Stenger, Bjorn and Lu, Tong},
  booktitle={Proceedings of the AAAI Conference on Artificial Intelligence},
  year={2023}
}

@article{Brateanu2025LYTNet,
  author={Brateanu, Alexandru and Balmez, Raul and Avram, Adrian and Orhei, Ciprian and Ancuti, Cosmin},
  journal={IEEE Signal Processing Letters},
  title={LYT-NET: Lightweight YUV Transformer-based Network for Low-Light Image Enhancement},
  year={2025},
  volume={32},
  pages={2065--2069},
  doi={10.1109/LSP.2025.3563125}
}

@inproceedings{Yan2025HVI,
  title={HVI: A New Color Space for Low-light Image Enhancement},
  author={Yan, Qingsen and Feng, Yixu and Zhang, Cheng and Pang, Guansong and Shi, Kangbiao and Wu, Peng and Dong, Wei and Sun, Jinqiu and Zhang, Yanning},
  booktitle={Proceedings of the IEEE/CVF Conference on Computer Vision and Pattern Recognition (CVPR)},
  year={2025}
}

@article{Jiang2021EnlightenGAN,
  title={EnlightenGAN: Deep Light Enhancement without Paired Supervision},
  author={Jiang, Yifan and Gong, Xinyu and Liu, Ding and Cheng, Yu and Fang, Chen and Shen, Xiaohui and Yang, Jianchao and Zhou, Pan and Wang, Zhangyang},
  journal={IEEE Transactions on Image Processing},
  volume={30},
  pages={2340--2349},
  year={2021},
  doi={10.1109/TIP.2021.3051462}
}

@inproceedings{Wang2022LLFlow,
  title={Low-Light Image Enhancement with Normalizing Flow},
  author={Wang, Yufei and Wan, Renjie and Yang, Wenhan and Li, Haoliang and Chau, Lap-Pui and Kot, Alex C.},
  booktitle={Proceedings of the AAAI Conference on Artificial Intelligence},
  volume={36},
  number={3},
  pages={2604--2612},
  year={2022}
}

@inproceedings{Weng2024MambaLLIE,
  title={MambaLLIE: Implicit Retinex-Aware Low Light Enhancement with Global-then-Local State Space},
  author={Weng, Jiangwei and Yan, Zhiqiang and Tai, Ying and Qian, Jianjun and Yang, Jian and Li, Jun},
  booktitle={Advances in Neural Information Processing Systems (NeurIPS)},
  volume={37},
  year={2024}
}

@inproceedings{Hou2023GSAD,
  title={Global Structure-Aware Diffusion Process for Low-light Image Enhancement},
  author={Hou, Jinhui and Zhu, Zhiyu and Hou, Junhui and Liu, Hui and Zeng, Huanqiang and Yuan, Hui},
  booktitle={Advances in Neural Information Processing Systems (NeurIPS)},
  volume={36},
  year={2023}
}

@article{Guo2017LIME,
  title={LIME: Low-Light Image Enhancement via Illumination Map Estimation},
  author={Guo, Xiaojie and Li, Yu and Ling, Haibin},
  journal={IEEE Transactions on Image Processing},
  volume={26},
  number={2},
  pages={982--993},
  year={2017},
  doi={10.1109/TIP.2016.2639450}
}

@article{Hu2023Bread,
  title={Low-light Image Enhancement via Breaking Down the Darkness},
  author={Hu, Qiming and Guo, Xiaojie},
  journal={International Journal of Computer Vision},
  year={2023}
}

@article{Zhou02012025,
author = {Libing Zhou and Xiaojing Chen and Baisong Ye and Xueli Jiang and Sheng Zou and Liang Ji and Zhengqian Yu and Jianjian Wei and Yexin Zhao and Tianyu Wang},
title = {A low-light image enhancement method based on HSV space},
journal = {The Imaging Science Journal},
volume = {73},
number = {1},
pages = {16--29},
year = {2025},
publisher = {Taylor \& Francis},
doi = {10.1080/13682199.2023.2266308},


URL = { 
    
        https://doi.org/10.1080/13682199.2023.2266308
    
    

},
eprint = { 
    
        https://doi.org/10.1080/13682199.2023.2266308
    
    

}

}

@inproceedings{Cui2022IAT,
  title={You Only Need 90K Parameters to Adapt Light: A Light Weight Transformer for Image Enhancement and Exposure Correction},
  author={Cui, Ziteng and Li, Kunchang and Gu, Lin and Su, Shenghan and Gao, Peng and Jiang, Zhengkai and Qiao, Yu and Harada, Tatsuya},
  booktitle={British Machine Vision Conference (BMVC)},
  year={2022}
}

@inproceedings{Fu2023PairLIE,
  title={Learning a Simple Low-Light Image Enhancer From Paired Low-Light Instances},
  author={Fu, Zhenqi and Yang, Yan and Tu, Xiaotong and Huang, Yue and Ding, Xinghao and Ma, Kai-Kuang},
  booktitle={Proceedings of the IEEE/CVF Conference on Computer Vision and Pattern Recognition (CVPR)},
  pages={22252--22261},
  year={2023}
}

@article{Weng2025CPGANet,
  title={A Lightweight Low-Light Image Enhancement Network via Channel Prior and Gamma Correction},
  author={Weng, Shyang-En and Miaou, Shaou-Gang and Christanto, Ricky},
  journal={International Journal of Pattern Recognition and Artificial Intelligence},
  volume={39},
  number={12},
  pages={2554013},
  year={2025},
  doi={10.1142/S0218001425540138}
}

@misc{Weng2025RTI,
  title={Rethinking Theoretical Illumination for Efficient Low-Light Image Enhancement},
  author={Weng, Shyang-En and Hsiao, Cheng-Yen and Lu, Li-Wei and Huang, Yu-Shen and Chen, Ting-Hao and Miaou, Shaou-Gang and Christanto, Ricky},
  year={2025},
  eprint={2409.05274},
  archivePrefix={arXiv},
  primaryClass={cs.CV}
}

@article{Ma2015MEFSSIM,
  author={Ma, Kede and Zeng, Kai and Wang, Zhou},
  title={Perceptual Quality Assessment for Multi-Exposure Image Fusion},
  journal={IEEE Transactions on Image Processing},
  volume={24},
  number={11},
  pages={3345--3356},
  year={2015},
  doi={10.1109/TIP.2015.2442920}
}

@article{Wang2013NPEA,
  author    = {Shuhang Wang and
               Jin Zheng and
               Hai-Miao Hu and
               Bo Li},
  title     = {Naturalness Preserved Enhancement Algorithm for Non-Uniform Illumination Images},
  journal   = {IEEE Transactions on Image Processing},
  volume    = {22},
  number    = {9},
  pages     = {3538--3548},
  year      = {2013},
  doi       = {10.1109/TIP.2013.2261309}
}

@article{Vonikakis2018Evaluation,
  author  = {Vonikakis, Vassilios and Kouskouridas, Rigas and Gasteratos, Antonios},
  title   = {On the Evaluation of Illumination Compensation Algorithms},
  journal = {Multimedia Tools and Applications},
  volume  = {77},
  number  = {8},
  pages    = {9211--9231},
  year     = {2018},
  month    = apr,
  doi      = {10.1007/s11042-017-4783-x}
}

@article{Lee2013LDR,
  author  = {Lee, Chulwoo and Lee, Chul and Kim, Chang-Su},
  title   = {Contrast Enhancement Based on Layered Difference Representation of 2D Histograms},
  journal = {IEEE Transactions on Image Processing},
  volume  = {22},
  number  = {12},
  pages   = {5372--5384},
  year    = {2013},
  month   = dec,
  doi     = {10.1109/TIP.2013.2284059}
}

@inproceedings{Brateanu2026Multinex,
  title={Multinex: Lightweight Low-light Image Enhancement via Multi-prior Retinex},
  author={Brateanu, Alexandru and Mu, Tingting and Ancuti, Codruta and Ancuti, Cosmin},
  booktitle={Proceedings of the IEEE/CVF Conference on Computer Vision and Pattern Recognition (CVPR)},
  year={2026}
}

@inproceedings{Park2019SPADE,
  title={Semantic Image Synthesis with Spatially-Adaptive Normalization},
  author={Park, Taesung and Liu, Ming-Yu and Wang, Ting-Chun and Zhu, Jun-Yan},
  booktitle={Proceedings of the IEEE/CVF Conference on Computer Vision and Pattern Recognition (CVPR)},
  pages={2337--2346},
  year={2019}
}

@inproceedings{han2024glare,
  title={Glare: Low light image enhancement via generative latent feature based codebook retrieval},
  author={Zhou, Han and Dong, Wei and Liu, Xiaohong and Liu, Shuaicheng and Min, Xiongkuo and Zhai, Guangtao and Chen, Jun},
  booktitle={European Conference on Computer Vision},
  pages={36--54},
  year={2024},
  organization={Springer}
}

@article{10.1109/TITS.2024.3359755,
author = {Qu, Jingxiang and Liu, Ryan Wen and Gao, Yuan and Guo, Yu and Zhu, Fenghua and Wang, Fei-Yue},
title = {Double Domain Guided Real-Time Low-Light Image Enhancement for Ultra-High-Definition Transportation Surveillance},
year = {2024},
issue_date = {Aug. 2024},
publisher = {IEEE Press},
volume = {25},
number = {8},
issn = {1524-9050},
url = {https://doi.org/10.1109/TITS.2024.3359755},
doi = {10.1109/TITS.2024.3359755},
journal = {Trans. Intell. Transport. Sys.},
month = aug,
pages = {9550–9562},
numpages = {13}
}


\end{document}